\documentclass[11pt]{article}
\usepackage{fullpage}
\usepackage{mathtools}
\usepackage{amsfonts}
\usepackage{amsthm}
\usepackage{hyperref}
\usepackage{xcolor}
\usepackage{enumerate}
\usepackage{indentfirst}
\usepackage{subfigure}
\usepackage[ruled,vlined,linesnumbered]{algorithm2e}
\hypersetup{
	colorlinks	= true,
	urlcolor	= orange,
	citecolor	= blue,
	linkcolor	= red
}

\title{A Theory of Selective Prediction}

\author{
    Mingda Qiao\\
    \texttt{mqiao@stanford.edu}
\and
    Gregory Valiant\\
    \texttt{valiant@stanford.edu}\thanks{This work is supported by NSF awards CCF-1704417 and AF:1813049 and by ONR award N00014-18-1-2295.}
}
\date{}

\newcommand{\1}[1]{\mathbb{I}\left[#1\right]}
\newcommand{\A}{\mathcal{A}}
\newcommand{\argmin}{\operatorname{argmin}}
\newcommand{\D}{\mathcal{D}}
\newcommand{\EMD}{\mathrm{EMD}}
\newcommand{\eps}{\epsilon}
\newcommand{\Ex}[1]{\mathbb{E}\left[#1\right]}
\newcommand{\I}{\mathcal{I}}
\newcommand{\J}{\mathcal{J}}
\renewcommand{\L}{\mathcal{L}}
\renewcommand{\over}[2]{\genfrac{}{}{0pt}{1}{#1}{#2}}
\newcommand{\R}{\mathbb{R}}
\newcommand{\td}{\text{d}} 
\newcommand{\U}{\mathcal{U}}
\newcommand{\var}[1]{\operatorname{Var}\left[#1\right]}
\newcommand{\X}{\mathcal{X}}

\newtheorem{theorem}{Theorem}[section]
\newtheorem{lemma}[theorem]{Lemma}
\newtheorem{definition}[theorem]{Definition}
\newtheorem{remark}[theorem]{Remark}
\newtheorem{proposition}[theorem]{Proposition}

\begin{document}

\maketitle

\begin{abstract}%
  We consider a model of \emph{selective prediction}, where the prediction algorithm is given a data sequence in an online fashion and asked to predict a pre-specified statistic of the upcoming data points. The algorithm is allowed to choose when to make the prediction as well as the length of the prediction window, possibly depending on the observations so far. We prove that, even without \emph{any} distributional assumption on the input data stream, a large family of statistics can be estimated to non-trivial accuracy. To give one concrete example, suppose that we are given access to an arbitrary binary sequence $x_1, \ldots, x_n$ of length $n$.  Our goal is to accurately predict the average observation, and we are allowed to choose the window over which the prediction is made: for some $t < n$ and $m \le n - t$, after seeing $t$ observations we predict the average of $x_{t+1}, \ldots, x_{t+m}$. This particular problem was first studied in~\cite{drucker2013high} and referred to as the ``density prediction game''. We show that the expected squared error of our prediction can be bounded by $O(\frac{1}{\log n})$ and prove a matching lower bound, which resolves an open question raised in~\cite{drucker2013high}. This result holds for any sequence (that is not adaptive to when the prediction is made, or the predicted value), and the expectation of the error is with respect to the randomness of the prediction algorithm. Our results apply to more general statistics of a sequence of observations, and we highlight several open directions for future work. 
\end{abstract}

\section{Introduction}
    Consider the following prediction problem: each day you observe the stock market, and at some point within the next $n$ days, you must make a prediction about the average return, or average volatility, of the stock market over some (future) period of time. Crucially, \emph{you get to choose both the timepoint within the $n$ days when you make the prediction, as well as the interval over which your prediction spans.} Without any distributional assumptions on the daily movements in the stock market, is it possible to accurately make such a prediction about the future?  As we show, the answer is ``yes'', and the expected error of the prediction tends to zero as $n$---the length of the window in which the prediction must occur---tends to infinity, assuming an absolute bound on the magnitude of daily fluctuations.  
    
    We consider several new angles to this age-old problem of making an accurate prediction about the future, given access to a sequence of observations.  The setting we consider abstracts three crucial properties of the above prediction problem: 1) We make no distributional assumptions about the sequence of observations. 2) The sequence, while possibly adversarial, is not adaptive, and is chosen independently of our prediction and when we make it. 3) We decide both when to make our prediction, as well as the duration over which our prediction spans (provided that both occur within some pre-specified horizon, denoted by $n$ in the example above). 
    
   In some sense, this model can be viewed as an exploration of the power that comes with being able to decide when to make a prediction about the future, in a world which, while possibly adversarial and changeable, is indifferent to your predictions (i.e. adversarial but non-adaptive). As such, it captures a number of important and natural online prediction tasks, beyond the toy example of stock-market predictions.

    A general formalization of this selective prediction problem can be framed as follows.  We are given a family of $n$ functions $(f_1, \ldots, f_n)$ where each $f_m: \X^m \to \R$. The prediction procedure proceeds as the following game. A sequence $x \in \X^n$ of length $n$ is chosen adversarially at the beginning of the game. The prediction game proceeds in $n$ rounds. At each time step $t \in \{0, \ldots, n - 1\}$, the player can make a claim in the following form: the function value of the next $m$ entries of the sequence ($1 \le m \le n - t$) is $\hat\alpha$. In this case, the game terminates immediately and the player incurs a loss of $\ell(\hat\alpha, \alpha)$, where $\alpha = f_m(x_{t+1}, \ldots, x_{t+m})$ is the actual function value on $x_{t+1}, \ldots, x_{t+m}$. Two natural loss functions that we focus on are the squared loss $\ell_2(\hat\alpha, \alpha) = \left(\hat\alpha - \alpha\right)^2$ and the absolute loss $\ell_1(\hat\alpha, \alpha) = \left|\hat\alpha - \alpha\right|$.
    If the player does not make a prediction at time $t$, the next data point $x_{t+1}$ is revealed to the player and the game continues. The player must predict exactly once before the data sequence is entirely observed.
    
    Facing an arbitrary and possibly adversarial data sequence, the predictor is only entitled the power of choosing the window over which the prediction is made. This power is indeed minimal in the sense that if the adversary knows in advance either the time step $t$ at which a prediction is made or the window length $m$, the predictor cannot achieve a non-trivial loss even for the task of predicting the arithmetic mean; see Section~\ref{sec:selectivity} for more details.

    This setting, and a related setting where one is must make a prediction about a single timestep, were first considered in~\cite{drucker2013high}.  These models deviate significantly from many other prediction settings, which typically either make strong distributional assumptions on the sequence of observations (e.g., that they are drawn independently, or generated from a Markov model, Hidden Markov Model, or exchangeable sequence, etc.), or make no assumptions but quantify the accuracy in terms of some notion of ``regret'' with respect to a limited set of benchmarks. Additionally, most previously studied prediction settings assume that the predictor must make a prediction at a specified time, or must make predictions at \emph{every} time step.  We discuss these differences, and connections to other settings more in Section~\ref{sec:related}.
    
    \subsection{Overview of Results}
    \paragraph{Estimating the arithmetic mean.} We first state our main results on the concrete task of predicting the average of a bounded real-valued sequence. 
    
    \begin{theorem}\label{thm:mean-estimation-informal}
        Suppose that $\X = [0, 1]$ and the function family $(f_m)$ is the arithmetic mean, i.e.,
        \[f_m(x_1, \ldots, x_m) = \frac{1}{m}\sum_{i=1}^{m}x_i.\]
        There exists a prediction algorithm that achieves an expected squared loss of $O(\frac{1}{\log n})$ on any sequence of length $n$. Moreover, this bound is tight: there is a distribution over sequences of length $n$ for which no algorithm can achieve an expected loss better than $\Omega(\frac{1}{\log n})$.
    \end{theorem}
    
    The upper bound of $O(\frac{1}{\log n})$ was first given in~\cite{drucker2013high}, and the matching lower bound resolves one of the main open questions posed in that work.  At an intuitive level, the mean estimation algorithm follows from the observation that a sequence cannot have a high variance on both small and large scales: if an adversary generates a uniformly random sequence in $\{0, 1\}^n$ in the hope that each single data point is hard to predict, the average of the whole sequence would concentrate around $\frac{1}{2}$ and thus be predictable. 
    
    The lower bound proof amounts to constructing a sequence with moderate variance at all different scales, simultaneously. Hence, no matter when the prediction algorithm chooses to make a prediction, and no matter the chosen time window, there will be a significant amount of variance in the values, conditioned on the sequence up to the time of prediction.  Consequently, the prediction algorithm has no hope in achieving too small a loss.
    
    \paragraph{Estimating smooth functions.} The positive result extends to other function families beyond the arithmetic mean. One such function family is the collection of all Lipschitz functions with respect to the earth mover's distance defined as follows. For a real sequence $(x_i)_{i=1}^{m}$ of length $m$, let $\U(x)$ denote the uniform distribution on the multiset $\{x_1, \ldots, x_m\}$, i.e., $\U(x)$ assigns probability mass $\frac{1}{m}\sum_{i=1}^{m}\1{x_i = x}$ to each $x$.
    
    \begin{definition}\label{def:EMD}
        The earth mover's distance $\EMD(x, y)$ between two real sequences $x$ and $y$ is defined as the Wasserstein distance between $\U(x)$ and $\U(y)$ with respect to the metric $d(a, b) = |a - b|$.
    \end{definition}
    
    \begin{definition}\label{def:smooth}
        A function $f: \R^m \to \R$ is $L$-smooth if and only if it is $L$-Lipschitz in earth mover's distance, i.e., $\left|f(x) - f(y)\right| \le L \cdot \EMD(x, y)$ for any $x, y \in \R^m$.
    \end{definition}
    
    We show that on bounded sequences, smooth functions can be estimated up to an absolute loss of $O\left(\frac{L}{\sqrt{\log n}}\right)$, where $L$ is the smoothness parameter and $n$ is the length of the input sequence.
    
    \begin{theorem}\label{thm:smooth-informal}
        Suppose $\X = [0, 1]$ and every function in $(f_m)$ is $L$-smooth. There exists a prediction algorithm that achieves an expected absolute loss of $O\left(\frac{L}{\sqrt{\log n}}\right)$ on any sequence of length $n$.
    \end{theorem}
    
    \paragraph{Estimating concatenation-concave functions.}
    In addition to the positive result on smooth functions, which only applies to functions on $\R^m$, we consider the following class of \emph{concatenation-concave} functions that admit a more general domain.
    
    \begin{definition}\label{def:concave}
        A function family $(f_m: \X^m \to \R)_{m=1}^{n}$ is concatenation-concave if and only if for any $x \in \X^{m_1}$ and $y \in \X^{m_2}$ with $m_1 + m_2 \le n$, it holds that
        \[
            f_{m_1+m_2}(x, y)
        \ge \frac{m_1}{m_1+m_2} f_{m_1}(x) + \frac{m_2}{m_1+m_2}f_{m_2}(y),
        \]
        where $f_{m_1+m_2}(x, y)$ is a shorthand for $f_{m_1+m_2}(x_1,\ldots,x_{m_1},y_1,\ldots,y_{m_2})$.
    \end{definition}
    
    Note that the arithmetic mean is concatenation-concave, with all inequalities in the above definition being equalities. Another family of concatenation-concave functions of practical importance is the following ``learnability'' function. Suppose that $\L$ is a given model class, which can be equivalently viewed as a family of bounded loss functions mapping $\X$ to $[0, 1]$. The learnability of a data sequence $(x_1, \ldots, x_m)$ is defined as $\inf_{\ell \in \L}\frac{1}{m}\sum_{i=1}^{m}\ell(x_i)$, the minimum average loss when we fit the sequence using a model in class $\L$.

    The learnability function is not captured by the family of smooth functions in the previous paragraph---in fact, $\X$ may not even be associated with a non-trivial metric. On the other hand, it can be easily verified that the learnability function is concatenation-concave.
    
    Our positive result for concatenation-concave functions states that any bounded concatenation-concave function can be estimated with an expected squared loss of $O(\frac{1}{\log n})$. This result is especially striking when considered in the context of estimating learnability, as the prediction accuracy is independent of the complexity of model class $\L$.
    
    \begin{theorem}\label{thm:concave-informal}
        Assuming that the function family $(f_m)$ is concatenation-concave and bounded in $[0, 1]$, there exists a prediction algorithm that achieves an expected squared loss of $O(\frac{1}{\log n})$ on any sequence of length $n$.
    \end{theorem}
    
    \paragraph{Fitting unseen data.} Given that we can accurately estimate the learnability of future data with respect to \emph{any} model class, it is natural to ask whether we can identify a model that actually fits the unseen data well. To this end, we consider the following generalization of our prediction model: instead of predicting $f_m(x_{t+1}, \ldots, x_{t+m})$, the predictor is required to output a model $\hat\ell$ in $\L$ that fits $x_{t+1}, \ldots, x_{t+m}$ well. The setting remains selective in the sense that $t$ and $m$ are still chosen by the prediction algorithm. The loss of the prediction is defined as the excess risk
        \[\frac{1}{m}\sum_{i=1}^{m}\hat\ell(x_{t+i}) - \inf_{\ell\in\L}\frac{1}{m}\sum_{i=1}^{m}\ell(x_{t+i}).\]
    By our results on mean estimation and a standard uniform convergence argument over $\L$, we can easily obtain an $O\left(\sqrt{\frac{|\L|}{\log n}}\right)$ upper bound on the optimal excess risk. Note that in this prediction task, the loss bound indeed depends on the cardinality of $\L$. In classic learning theory, however, the dependence of the excess risk on $|\L|$ is typically logarithmic. It remains a compelling open question whether the excess risk can be further improved to $O\left(\sqrt{\frac{\log|\L|}{\log n}}\right)$ as classical learning theory suggests, or whether a polynomial dependence on $|\L|$ is inevitable in the worst case.
    
    \subsection{Related Work}\label{sec:related}
    Most closely related to this paper is the work of \cite{drucker2013high}, which studies several prediction problems in the setting where we are given access to an arbitrary (adversarial) infinite binary sequence, and attempt to predict the value of a single index, or predict the fraction of 1's in a future interval.  Crucially, the predictor is also allowed to choose the prediction window selectively. \cite{drucker2013high} shows that given a horizon of length $2^{O(1/\eps)}$, one can achieve a squared error of at most $\eps$ in expectation, which translates into an expected squared loss of $O(\frac{1}{\log n})$ in our setting. Our work recovers this result as a special case, and proves a matching lower bound which implies that this exponential dependence on $1/\eps$ is necessary.
    
    The recent work \cite{feige2017chasing} proves a \emph{local repetition lemma}, which states that a sufficiently long sequence must exhibit a certain level of pattern at some time scale. The difference from our work is the interpretation of this observation: while \cite{feige2017chasing} addresses the online learning setting where the regret is defined with respect to a set of ``stateful'' policies that can be represented by state machines, we consider the problem of directly predicting an arbitrary sequence and aim to generalize this observation to a broader class of prediction and learning tasks.
    
    More broadly, sequential prediction and decision making is a major subject of research in many different fields. Early study on this problem dates back to the pioneering work of~\cite{hannan1957approximation} in the 1950s. This problem, along with many of its extensions, is addressed under various terminologies in different communities, including ``universal prediction'' in information theory~\cite{feder1992universal}, ``universal portfolios'' in mathematical finance~\cite{cover1991universal,cover1996universal,blum1999universal} and ``online learning'' in machine learning theory~\cite{littlestone1994weighted,cesa1997use,cesa2006prediction}. In particular, our approach is closely related to yet different from the online learning formulation. In online learning, the predictor has access to a class of strategies (also known as ``experts''). The prediction algorithm leverages the expert advice and makes sequential prediction on \emph{every} time step. The performance of the predictor is measured in terms of the regret, defined as the difference between the incurred loss and the loss of the best expert in hindsight.    
    
    There is also a large body of work on ``conformal prediction'' in the online setting where datapoints are revealed one at a time (see e.g.\ the book~\cite{vovk2005conformal}).  This body of work is largely concerned with understanding how confidently one can make a prediction about the label, $y_t$, given $x_t$ and a sequence of labeled data $(x_1,y_1),\ldots, (x_{t-1},y_{t-1})$.  In general, strong positive results exist in the independent setting where data is drawn independently from a fixed distribution, and also in the more general setting where the sequence of data is assumed to be exchangeable.  
    
    The selective prediction model we consider is significantly different from the above two settings.  In contrast to the regret minimization framework, we do not restrict ourselves to a specified family of experts; instead, we evaluate the predictor solely based on the expected loss rather than the loss relative to the best expert.  In contrast to work on conformal prediction, our results hold without any distributional assumptions on the sequence of data.  Crucially, to enable these strong results, in our model the prediction algorithm is allowed to be \emph{selective} in the sense that its prediction may not necessarily cover the entire time horizon, and the prediction can be made over an interval of arbitrary length instead of a single observation.   
    
    We note that the recent work of~\cite{sharan2018prediction} addresses the problem of predicting the distribution of the next observation in the data sequence from a different perspective. The focus of their work is whether accurate prediction can be made using a small memory, and their results apply to the scenario where the data stream is drawn from a distribution with bounded mutual information between the past and the future (for example, a sequence generated by a hidden Markov model). In contrast, our model captures the prediction of a more general family of statistics of the upcoming observations, and we make no distributional assumptions on the sequence.

    Another related line of research concerns the estimation of learnability given limited data. In more detail, given labeled data drawn i.i.d.\ from an underlying distribution, we are asked to estimate how well a given model class can fit the distribution. It is shown that for linear models, a sample of size $O(\sqrt{d})$ is sufficient for accurate estimation~\cite{dicker2014variance,kong2018estimating}, and this is much less than the amount of data needed to learn a linear model. Our work is incomparable to this line of research, since our results apply to the more general setting where the data are not assumed to be i.i.d.\ and the model class $\L$ can be arbitrary.

\section{Tight Loss Bounds for Mean Estimation}
    We start by studying a special case of the general prediction problem: estimating the mean of a bounded sequence. Without loss of generality, we assume that the instance space is $\X = [0, 1]$. The function value on a subsequence of numbers is simply the arithmetic mean, i.e., $f_m(x_1, \ldots, x_m) = \frac{1}{m}\sum_{i=1}^{m}x_i$.
    
	\subsection{Selective Predictor with Vanishing Loss}
	We begin by presenting the simple prediction scheme from~\cite{drucker2013high} that achieves an error which goes to zero as $n$ tends to infinity, and include a slightly simpler proof of the $O(\frac{1}{\log n})$ loss.  In the following, we assume that the sequence length $n$ is a power of two. Let $\U(S)$ denote the uniform distribution over the finite set $S$.
	
	\begin{algorithm}[ht]
	    \caption{Selective Prediction} \label{alg:selective-prediction}
	    \KwIn{Sequence $x \in \X^n$ of length $n = 2^k$.}
	    $k' \gets  \U([k])$\;
	    $t \gets \U(\{0, 2^{k'}, 2\cdot 2^{k'}, \ldots, n - 2^{k'}\})$\;
	    Observe $x_1, x_2, \ldots, x_{t+2^{k'-1}}$\;
	    $\hat\alpha \gets f_{2^{k'-1}}(x_{t+1}, \ldots, x_{t+2^{k'-1}})$\;
	    Predict that $f_{2^{k'-1}}(x_{t+2^{k'-1}+1}, \ldots, x_{t+2^{k'}})$ equals $\hat\alpha$\;
	\end{algorithm}
	
	Algorithm~\ref{alg:selective-prediction} chooses the prediction window by drawing $k'$ and $t$ randomly at the beginning. Then, at time $t + 2^{k' - 1}$, the algorithm predicts that the average of the next $2^{k'-1}$ numbers is close to that of the most recent $2^{k'-1}$ numbers. We prove in the following that Algorithm~\ref{alg:selective-prediction} achieves a squared loss of $O(\frac{1}{\log n})$.
	
	\begin{lemma}\label{lem:mean-estimation}
	    Suppose that the instance space is $\X = [0, 1]$ and the function family $f$ is the arithmetic mean.
	    For any integer $k \ge 1$, Algorithm~\ref{alg:selective-prediction} achieves an expected squared loss of at most $\frac{1}{k}$ on any sequence of length $2^k$.
	\end{lemma}
	
	\begin{remark}\label{rem:power-of-two}
	Lemma~\ref{lem:mean-estimation} directly implies that $O(\frac{1}{\log n})$ squared loss can be achieved in the general case that $n$ is not a power of two, thus proving the upper bound part of Theorem~\ref{thm:mean-estimation-informal}. Indeed, choosing $k = \left\lfloor \log_2 n\right\rfloor$ and running Algorithm~\ref{alg:selective-prediction} as if the sequence is of length $2^k$ gives an expected squared loss of at most
	    $\frac{1}{\left\lfloor \log_2 n\right\rfloor} = O(\frac{1}{\log n})$.
	\end{remark}

    \begin{proof}
    For integer $k \ge 1$ and $\mu \in [0, 1]$, let $L(k, \mu)$ denote the maximum expected squared loss that Algorithm~\ref{alg:selective-prediction} incurs on a sequence of $2^k$ numbers between $0$ and $1$ with average $\mu$. We prove by induction on $k$ that $L(k, \mu) \le \frac{4\mu(1-\mu)}{k}$, which directly implies the proposition.
    
    When $k = 1$, Algorithm~\ref{alg:selective-prediction} reduces to predicting that $x_2 = x_1$, and the squared loss can be bounded as follows:
	\[
	    L(1, \mu)
	=   \sup_{\over{x_1, x_2 \in [0, 1]}{x_1 + x_2 = 2\mu}}(x_1 - x_2)^2
	=   \min(4\mu^2, 4(1-\mu)^2)
	\le 4\mu(1-\mu).
	\]
	
	For $k \ge 2$, we note that with probability $\frac{1}{k}$, Algorithm~\ref{alg:selective-prediction} chooses $k' = k$ and predicts that the last $2^{k-1}$ numbers have the same average as the first $2^{k-1}$ numbers. Let $\mu_1$ and $\mu_2$ denote the averages of the first and last $2^{k-1}$ numbers, respectively. Then, the squared loss in in this case is given by $(\mu_1 - \mu_2)^2$.
	With probability $\frac{k-1}{k}$, the algorithm chooses some $k' < k$ and the algorithm is equivalent to running the same algorithm either on either the first $2^{k-1}$ numbers or the last $2^{k-1}$ numbers. By the induction hypothesis, the conditional expected squared loss is upper bounded by
	    \[\frac{L(k-1, \mu_1) + L(k-1, \mu_2)}{2} \le \frac{2\mu_1(1-\mu_1) + 2\mu_2(1-\mu_2)}{k-1}.\]
	Based on the above analysis, we have
	\begin{align*}
	    L(k, \mu)
	&\le \sup_{\over{\mu_1, \mu_2 \in [0, 1]}{\mu_1+\mu_2=2\mu}}\left[\frac{1}{k}\cdot(\mu_1 - \mu_2)^2 + \frac{k-1}{k}\cdot\frac{2\mu_1(1-\mu_1) + 2\mu_2(1-\mu_2)}{k-1}\right]\\
	&= \frac{1}{k}\cdot\sup_{\over{\mu_1, \mu_2 \in [0, 1]}{\mu_1+\mu_2=2\mu}}\left[2(\mu_1 + \mu_2) - (\mu_1 + \mu_2)^2\right]
	= \frac{4\mu(1-\mu)}{k},
	\end{align*}
	which completes the proof.
	\end{proof}

	\subsection{Selectivity is Necessary}\label{sec:selectivity}
	Algorithm~\ref{alg:selective-prediction} is selective in the sense that it randomly chooses the time step $t$ as well as the window length $m$ for its prediction. Such selectivity is crucial to achieving a sub-constant loss. Intuitively, if $t$ is known to the adversary, the data stream can be chosen such that the first $t$ elements are independent of the rest, rendering any meaningful prediction unfeasible. Likewise, if the prediction window is of fixed length $m$, the data sequence can be constructed as blocks of size $m/2$, which also leads to a constant lower bound on the prediction loss. Finally, if the time, $t$, of the prediction can be chosen, but the window must contain the remaining $n-t$ observations, a constant lower bound also exists.  The formal proof of the following proposition is deferred to Appendix~\ref{app:selectivity}.
	
	\begin{proposition}\label{prop:selectivity}
	    Suppose that prediction algorithm $\A$, when running on a sequence of length $n$, either: (1) always predicts at the same time $t$, (2) always chooses the same window length $m$, or (3) chooses $t$, but must make a prediction over the entire window of $n-t$ remaining timesteps. Then, there exists a binary sequence of length $n$ on which $\A$ incurs an expected squared loss of at least $\frac{1}{64}$.
	\end{proposition}
	
	\subsection{Matching Lower Bound}
	The prediction scheme in Algorithm~\ref{alg:selective-prediction} may appear not to leverage all the power of the predictor; indeed, the algorithm chooses the prediction window at the beginning of the algorithm, while the model in general allows the algorithm to make the decision adaptively. Nevertheless, we show in the following that such adaptivity brings little marginal gain---the upper bound in Lemma~\ref{lem:mean-estimation} is optimal up to a constant factor.
	
	The key in our lower bound proof is to construct a sequence that simultaneously satisfies an anti-concentration property on both small and large timescales. Such a sequence guarantees that even after the predictor observes a prefix of the sequence, the average of the future data sequence still has a large conditional variance given the prefix. This implies a lower bound on the expected squared error achievable by any prediction algorithm.
	
\begin{figure}[ht]
	    \centering
	    \includegraphics[scale=0.5]{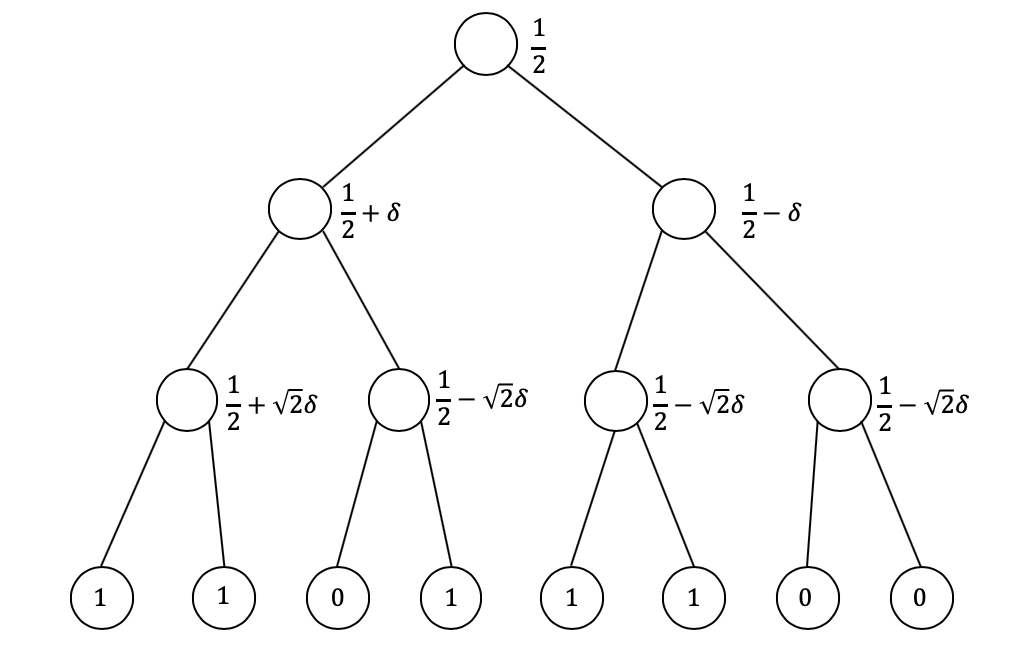}
	    \caption{Sample construction for $k = 3$, where $\delta = \frac{1}{2\sqrt{3}}$.}
	    \label{fig:lb-small}
	\end{figure}

\begin{figure}[ht]
	    \centering
	    \subfigure[Window length $2^{10}$]{
    	    \includegraphics[scale=0.35]{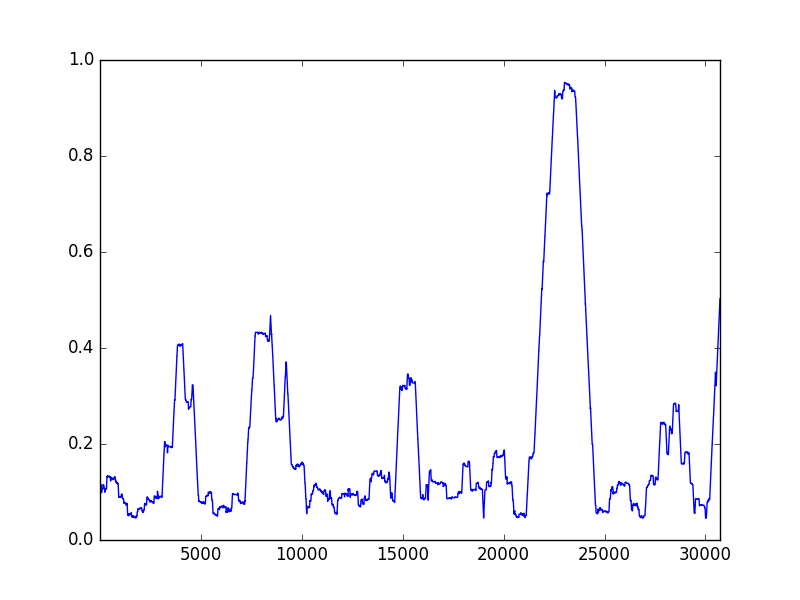}
	    }
	    \subfigure[Window length $2^{15}$]{
    	    \includegraphics[scale=0.35]{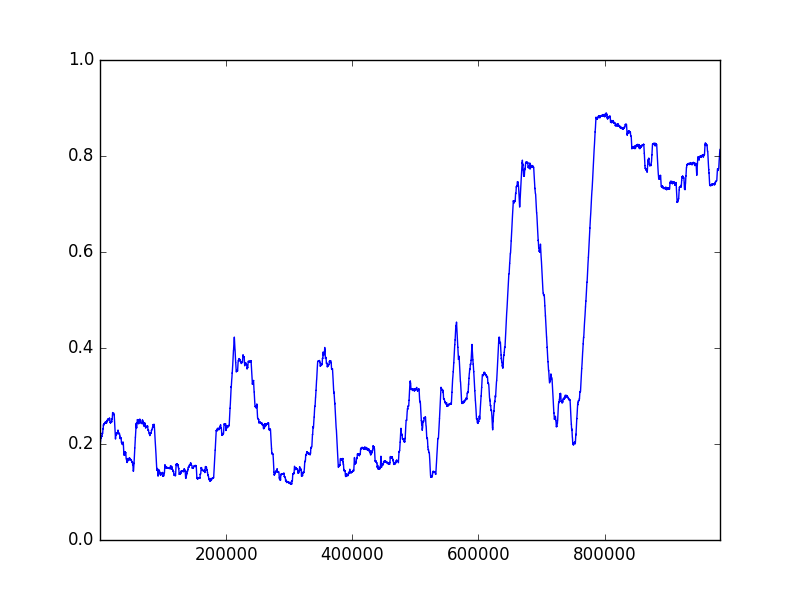}
	    }
	    \caption{Depiction of our lower bound construction for $k = 20$. For one sequence constructed according to our construction, the two plots each depict the average value in a moving window of a certain length, $w$, over the first $30$ (non-overlapping) windows of size $w$. In the left plot, the window length is $w=2^{10}$ and in the right plot, $w=2^{15}$. The fact that both plots look like similar stochastic processes, and in particular exhibit similar anti-concentration behaviors illustrates the main property of our construction: a single underlying sequence that is anti-concentrated, simultaneously, at all timescales.
	    }
	    \label{fig:lb-large}
	\end{figure}

	Again, we focus on the case that $n = 2^k$ is a power of two, as the proof can be extended to the general case (losing at most a constant factor) by the same argument as in Remark~\ref{rem:power-of-two}. Consider a perfect binary tree with $n$ leaves. In the following, we assign a real value between $0$ and $1$ to each node in the tree recursively, and the sequence $x \in [0, 1]^n$ is chosen as the values on the $n$ leaves. Let $\delta = \frac{1}{2\sqrt{k}}$. The value of the root is defined as $\frac{1}{2}$. Then, for each node at the $j$-th level of the tree (the root being at level $0$ and leaves at level $k$), we choose its value randomly and independently from $\frac{1}{2}\pm\sqrt{j}\cdot\delta$ such that the expectation of the value equals the value of its parent. In particular, if the parent has value $\frac{1}{2} + \sqrt{j-1}\cdot\delta$, the node takes value
	\[
	\begin{cases}
	\frac{1}{2} + \sqrt{j}\cdot\delta, & \text{with probability } \frac{\sqrt{j}+\sqrt{j-1}}{2\sqrt{j}},\\
	\frac{1}{2} - \sqrt{j}\cdot\delta, & \text{with probability } \frac{\sqrt{j}-\sqrt{j-1}}{2\sqrt{j}};\\
	\end{cases}
	\]
	the probabilities are switched in the other case. Note that by our choice of $\delta$, all leaves will be assigned values in $[0, 1]$ and thus the resulting sequence is bounded. See Figure~\ref{fig:lb-small} for a realization of the construction when $k = 3$. Also see Figure~\ref{fig:lb-large} for plots of a sample sequence for $k = 20$. Note that after taking the moving average at different scales, the sequence still exhibits strong anti-concentration. (In contrast, the moving average of a uniformly random bit string would concentrate around $\frac{1}{2}$ at larger scales.)
	
    Let $\D_n$ denote the distribution of the sequence that we defined as above. We show that any algorithm will incur an $\Omega(\frac{1}{\log n})$ squared loss in expectation given a random sequence drawn from $\D_n$. By an averaging argument, there exists a sequence on which the algorithm incurs an $\Omega(\frac{1}{\log n})$ squared loss. This proves the lower bound part of Theorem~\ref{thm:mean-estimation-informal}.

	\begin{lemma}\label{lem:estimation-lb}
	    For any integer $k \ge 1$ and $n = 2^k$, any prediction algorithm for the arithmetic mean incurs an expected squared loss of at least $\frac{1}{64k}$ on a random sequence drawn from $\D_n$.
	\end{lemma}
	
	\begin{remark}
	    Since the arithmetic mean is both $1$-smooth (Definition~\ref{def:smooth}) and concatenation-concave (Definition~\ref{def:concave}), Lemma~\ref{lem:estimation-lb} implies that an $\Omega(\frac{1}{\log n})$ squared loss is inevitable for these two function families.
	\end{remark}
	
	\begin{proof}
	    We show that for any $t \in \{0, \ldots, n - 1\}$ and $m \in [n - t]$, conditioned on any prefix $x_1, \ldots, x_t$ of the sequence, the variance in
	        $\frac{1}{m}\sum_{i=1}^{m}x_{t+i}$
	    is at least $\frac{1}{64k}$. The theorem follows from the observation that this is the smallest expected squared loss that can be achieved when the player decides to predict the average of $x_{t+1}, \ldots, x_{t+m}$.
	    
	    Fix $t$ and $m$. By our construction of $\D_n$, there exists integer $k'$ such that $2^{k'} \ge \frac{m}{4}$ and $(x_{t+1}, \ldots, x_{t+m})$ contains a contiguous subsequence $(x_{t'+1}, \ldots, x_{t' + 2^{k'}})$ of length $2^{k'}$ that exactly corresponds to the $2^{k'}$ leaves in a subtree of height $k'$. Let $u$ denote the root of the subtree. We can actually prove a stronger claim: the variance in $\frac{1}{m}\sum_{i=1}^{m}x_{t+i}$ is lower bounded by $\frac{1}{64k}$, even when conditioned on the values of \emph{all} nodes in the binary tree except the subtree rooted at $u$.

	    Let $v$ be the parent of $u$. Let $p_u$ and $p_v$ denote the values of $u$ and $v$ respectively. It can be verified from the construction that $\var{p_u|p_v} = \delta^2 = \frac{1}{4k}$.
	    Since the subtree rooted at $u$ has $2^{k'} \ge \frac{m}{4}$ leaves, the value of node $u$ contributes at least a $\frac{1}{4}$ fraction to the average $\frac{1}{m}\sum_{i=1}^{m}x_{t+i}$. It follows that the conditional variance in $\frac{1}{m}\sum_{i=1}^{m}x_{t+i}$ is lower bounded by $\left(\frac{1}{4}\right)^2\cdot\var{p_u|p_v} = \frac{1}{64k}$.
	\end{proof}

\section{Estimating General Functions}

    We extend the positive results for mean estimation to more general function families. It turns out that Algorithm~\ref{alg:selective-prediction} has a stronger guarantee beyond mean estimation: we will show that exactly the same algorithm also achieves a vanishing loss on smooth functions and concatenation-concave functions.

    \subsection{Smooth Functions}
    Recall that Algorithm~\ref{alg:selective-prediction} chooses $k'$ and $t$ randomly, and then uses
    $f_{2^{k'-1}}(x_{t+1}, \ldots, x_{t+2^{k'-1}})$
    as an estimate for
    $f_{2^{k'-1}}(x_{t+2^{k'-1}+1}, \ldots, x_{t+2^{k'}})$. We show in the following that the sequences $x_{t+1}, \ldots, x_{t+2^{k'-1}}$ and $x_{t+2^{k'-1}+1}, \ldots, x_{t+2^{k'}}$ are close in earth mover's distance defined as in Definition~\ref{def:EMD}. The prediction loss can then be bounded using the smoothness of $f$.
    
    \begin{lemma}\label{lem:smooth}
        Suppose that $\X = [0, 1]$ and every function in $(f_m)$ is $L$-smooth. For any integer $k \ge 1$, Algorithm~\ref{alg:selective-prediction} achieves an expected absolute loss of at most $\frac{L}{\sqrt{k}}$ on any sequence of length $2^k$.
    \end{lemma}
    
    Lemma~\ref{lem:smooth} implies Theorem~\ref{thm:smooth-informal} by the argument in Remark~\ref{rem:power-of-two}.
    
    \begin{proof}
        Let $\I_{k',t}$ and $\J_{k',t}$ denote subsequences $x_{t+1}, \ldots, x_{t+2^{k'-1}}$ and $x_{t+2^{k'-1}+1}, \ldots, x_{t+2^{k'}}$. In the following, we prove the an upper bound on the expected earth mover's distance between $\I_{k',t}$ and $\J_{k',t}$:
        \[\Ex{\EMD(\I_{k',t}, \J_{k',t})} \le \frac{1}{\sqrt{k}},\]
        where the expectation is taken over the randomness in $k'$ and $t$.
        
        It is well-known that the earth mover's distance between two distributions on $[0, 1]$ can be rewritten as
        \[
            \EMD(\I_{k',t}, \J_{k',t})
        =   \int_{0}^{1}\left|\U(\I_{k',t})([0, \tau]) - \U(\J_{k',t})([0, \tau])\right|~\td \tau.
        \]
        Recall that $\U(\I_{k',t})$ (resp. $\U(\J_{k',t})$) denotes the uniform distributions naturally defined by $\I_{k',t}$ (resp. $\J_{k',t}$), i.e.,
        $\U(\I_{k',t})([0, \tau]) = \frac{1}{2^{k'-1}}\sum_{i=1}^{2^{k'-1}}\1{x_{t+i} \in [0, \tau]}$.
        
        Fix $\tau \in [0, 1]$ and consider an auxiliary sequence $x^{(\tau)}$ defined as follows:
		    \[x^{(\tau)}_i = \1{x_i \in [0, \tau]}.\]
	    Then, $\U(\I_{k',t})([0, \tau])$ and $\U(\J_{k',t})([0, \tau])$ are exactly the means of subsequences $x^{(\tau)}_{t+1},\ldots,x^{(\tau)}_{t+2^{k'-1}}$ and $x^{(\tau)}_{t+2^{k'-1}+1},\ldots,x^{(\tau)}_{t+2^{k'}}$, respectively. Since $x^{(\tau)}$ is bounded in $[0, 1]$, by Lemma~\ref{lem:mean-estimation},
	    \begin{align*}
            &\Ex{\left|\U(\I_{k',t})([0, \tau]) - \U(\J_{k',t})([0, \tau])\right|}\\
        \le &\sqrt{\Ex{\left(\U(\I_{k',t})([0, \tau]) - \U(\J_{k',t})([0, \tau])\right)^2}}\tag{concavity of $\sqrt{x}$}\\
        \le &\frac{1}{\sqrt{k}}.\tag{Lemma~\ref{lem:mean-estimation}}
	    \end{align*}

	    Taking an integral over $\tau \in [0, 1]$ proves that
	    \[
	        \Ex{\EMD(\I_{k',t}, \J_{k',t})}
	    = \int_{0}^{1}\Ex{\left|\U(\I_{k',t})([0, \tau]) - \U(\J_{k',t})([0, \tau])\right|}~\td \tau
	    \le\frac{1}{\sqrt{k}},
	    \]
	    which completes the proof, since the expected absolute loss is upper bounded by
        \[
            \Ex{\left|f_{2^{k'-1}}(\I_{k',t})-f_{2^{k'-1}}(\J_{k',t})\right|}
        \le\Ex{L\cdot\EMD(\I_{k',t}, \J_{k',t})}
        \le\frac{L}{\sqrt{k}}
        \]
        due to the $L$-smoothness of $(f_m)$.
    \end{proof}

    \subsection{Concatenation-Concave Functions}
    Algorithm~\ref{alg:selective-prediction} also applies to the case where the function family to be predicted is concatenation-concave. The proof resembles that of Lemma~\ref{lem:mean-estimation}, yet a slightly different induction hypothesis is used. Again, Lemma~\ref{lem:concave} readily extends to the general case where the sequence length is not a power of two and thus proves Theorem~\ref{thm:concave-informal}.
    
    \begin{lemma}\label{lem:concave}
        Suppose that the function family $(f_m)$ is concatenation-concave and bounded in $[0, 1]$. For any integer $k \ge 1$, Algorithm~\ref{alg:selective-prediction} achieves an expected squared loss of at most $\frac{4}{k}$ on any sequence of length $2^k$.
    \end{lemma}
    
    \begin{proof}
	For integer $k \ge 1$ and $\mu \in [0, 1]$, let $L(k, \mu)$ denote the maximum expected squared loss that Algorithm~\ref{alg:selective-prediction} incurs on a sequence of length $2^k$ with function value $f_{2^k}(x_1, \ldots, x_{2^k}) = \mu$. Let $\mu_1 = f_{2^{k-1}}(x_1, \ldots, x_{2^{k-1}})$ and $\mu_2 = f_{2^{k-1}}(x_{2^{k-1}+1}, \ldots, x_{2^k})$.
	By the concatenation-concavity of $(f_m)$, we have $\mu_1 + \mu_2 \le 2\mu$. In the following, we prove by induction that
		$L(k, \mu) \le \frac{4\mu(2-\mu)}{k}$,
	which further implies that $L(k, \mu) \le \frac{4}{k}$ for any $\mu \in [0, 1]$.

	When $k = 1$, the squared loss is upper bounded by
	\[
	    L(1, \mu)
	=   \max_{\over{\mu_1, \mu_2 \in [0, 1]}{\mu_1 + \mu_2 \le 2\mu}}(\mu_1 - \mu_2)^2
	\le \min(4\mu^2, 4(1 - \mu)^2) \le 4\mu(2 - \mu).
	\]

	Suppose that $k \ge 2$. With probability $\frac{1}{k}$, Algorithm~\ref{alg:selective-prediction} chooses $k' = k$ and the loss is given by $(\mu_1 - \mu_2)^2$. With probability $\frac{k-1}{k}$, the algorithm chooses $k' \ne k$, and the algorithm is equivalent to running the same algorithm on either the first or last $2^{k-1}$ entries of the sequence. The conditional expected loss in this case is upper bounded, thanks to the induction hypothesis, by
	\[
	    \frac{L(k-1, \mu_1) + L(k-1, \mu_2)}{2}
	\le \frac{2\mu_1(2-\mu_1)+2\mu_2(2-\mu_2)}{k-1}.
	\]
	To sum up, we have
	\begin{align*}
		L(k, \mu)
	\le	&\sup_{\over{\mu_1, \mu_2 \in [0, 1]}{\mu_1 + \mu_2 \le 2\mu}}\left[\frac{1}{k}(\mu_1 - \mu_2)^2 + \frac{k-1}{k}\cdot\frac{2\mu_1(2-\mu_1)+2\mu_2(2-\mu_2)}{k-1}\right]\\
	=	&\frac{1}{k}\cdot\sup_{\over{\mu_1, \mu_2 \in [0, 1]}{\mu_1 + \mu_2 \le 2\mu}}\left[4(\mu_1+\mu_2)-(\mu_1+\mu_2)^2\right]
	=	\frac{4\mu(2-\mu)}{k}
	\end{align*}
	as desired, where the last step follows from $0\le\mu_1+\mu_2\le 2\mu \le 2$ and the monotonicity of $4x-x^2$ on $[0, 2]$.
	\end{proof}

\section{Fitting Unseen Data}
    In this section, we study the problem of finding a model that fits the upcoming data points with a small excess risk. We consider a finite model class $\L$, each element of which can be viewed as a loss function $\ell: \X \to [0, 1]$. The goal of the player is to choose some time step $t$ and window length $m$ and output a model $\hat\ell$ that minimizes the excess risk defined as follows:
        \[\frac{1}{m}\sum_{i=1}^{m}\hat\ell(x_{t+i}) - \inf_{\ell \in \L}\frac{1}{m}\sum_{i=1}^{m}\ell(x_{t+i}).\]
    A natural approach to this problem is to follow the strategy in Algorithm~\ref{alg:selective-prediction} and output the ``empirical risk minimizer'' (ERM) of observed data. We formally state the algorithm as follows. The excess risk of Algorithm~\ref{alg:ERM} can be bounded by a uniform convergence argument over all models in $\L$.

	\begin{algorithm}[ht]
	    \caption{Empirical Risk Minimization} \label{alg:ERM}
	    \KwIn{Model class $\L$ and sequence $x \in \X^n$ of length $n = 2^k$.}
	    $k' \gets  \U([k])$\;
	    $t \gets \U(\{0, 2^{k'}, 2\cdot 2^{k'}, \ldots, n - 2^{k'}\})$\;
	    Observe $x_1, x_2, \ldots, x_{t+2^{k'-1}}$\;
	    $\hat\ell \gets \argmin_{\ell \in \L}\frac{1}{m}\sum_{i=1}^{m}\ell(x_{t+i})$\;
	    Predict that $\hat\ell$ minimizes the risk on $(x_{t+2^{k'-1}+1}, \ldots, x_{t+2^{k'}})$\;
	\end{algorithm}

    \begin{proposition}\label{prop:ERM-upper}
        For any integer $k \ge 1$ and finite model class $\L$, Algorithm~\ref{alg:ERM} achieves an expected excess risk of at most $O\left(\sqrt{\frac{|\L|}{k}}\right)$ on any sequence of length $2^k$.
    \end{proposition}
    
    \begin{proof}
        Let $\I_{k',t}$ and $\J_{k',t}$ denote sequences $(x_{t+1},\ldots,x_{t+2^{k'-1}})$ and $(x_{t+2^{k'-1}+1},\ldots,x_{t+2^{k'}})$. For $\ell \in \L$, let $\ell(\I_{k',t})$ denote the average loss of $\ell$ on sequence $\I_{k',t}$. By a standard uniform convergence argument, the expected excess risk of Algorithm~\ref{alg:ERM} is upper bounded by
        \begin{align*}
            &\Ex{2\max_{\ell \in \L}\left|\ell(\I_{k',t})-\ell(\J_{k',t})\right|}\\
        =  ~&2\Ex{\sqrt{\max_{\ell \in \L}\left[\ell(\I_{k',t})-\ell(\J_{k',t})\right]^2}}\\
        \le~&2\sqrt{\Ex{\max_{\ell \in \L}\left[\ell(\I_{k',t})-\ell(\J_{k',t})\right]^2}} \tag{concavity of $\sqrt{x}$}\\
        \le~&2\sqrt{\sum_{\ell \in \L}\Ex{\left(\ell(\I_{k',t})-\ell(\J_{k',t})\right)^2}}
        = O\left(\sqrt{\frac{|\L|}{k}}\right). \tag{Lemma~\ref{lem:mean-estimation}}
        \end{align*}
    \end{proof}
    
    Falling short of proving a lower bound that matches Proposition~\ref{prop:ERM-upper}, we show that further improving the excess risk would require a more sophisticated prediction scheme than Algorithm~\ref{alg:ERM}. In particular, Proposition~\ref{prop:ERM-lower} states that when $\left|\L\right| = \Theta(\log n)$, Algorithm~\ref{alg:ERM} incurs a constant excess risk in expectation and thus the upper bound in Proposition~\ref{prop:ERM-upper} is almost tight for Algorithm~\ref{alg:ERM}.
    
    \begin{proposition}\label{prop:ERM-lower}
        For any integer $k \ge 2$, there exists a model class $\L$ of size $k$ and a sequence $(x_t)$ of length $2^k$ such that Algorithm~\ref{alg:ERM} incurs an expected excess risk of at least $\frac{1}{8}$ on $(x_t)$.
    \end{proposition}
    
    \begin{proof}
        Let $\X = \{0, 1, \ldots, 2^k-1\}$ and $\L = \{\ell_1, \ell_2, \ldots, \ell_k\}$. Each $\ell_i(x)$ is defined as:
        \[
            \ell_i(x)
        =   \begin{cases}
            1, & \left\lfloor \frac{x}{2^{i-1}} \right\rfloor \text{ is odd,}\\
            i\cdot\eps, & \text{otherwise,}
        \end{cases}
        \]
        where $\eps = \frac{1}{4k}$. The input sequence is defined as $x_t = t - 1$. An example of the construction with $k = 3$ is shown in Table~\ref{tab:construction}.
        
        \begin{table}[ht]
            \centering
            \begin{tabular}{|c|c|c|c|c|c|c|c|c|}
            \hline
                $x_t$           & 0 & 1 & 2 & 3 & 4 & 5 & 6 & 7\\
            \hline
                $\ell_1(x_t)$   & $\eps$ & 1 & $\eps$ & 1 & $\eps$ & 1 & $\eps$ & 1 \\
            \hline
                $\ell_2(x_t)$   & $2\eps$ & $2\eps$ & 1 & 1 & $2\eps$ & $2\eps$ & 1 & 1 \\
            \hline
                $\ell_3(x_t)$   & $3\eps$ & $3\eps$ & $3\eps$ & $3\eps$ & 1 & 1 & 1 & 1 \\
            \hline
            \end{tabular}
            \caption{Construction for $k=3$.}
            \label{tab:construction}
        \end{table}
        
        Let $\I_{k',t}$ and $\J_{k',t}$ denote subsequences $x_{t+1},\ldots,x_{t+2^{k'-1}}$ and $x_{t+2^{k'-1}+1},\ldots,x_{t+2^{k'}}$. For $\ell \in \L$, let $\ell(\I_{k',t})$ denote the average loss of $\ell$ on sequence $\I_{k',t}$. It can be verified that
        \[
            \ell_{i}(\I_{k',t})
        =   \begin{cases}
            \frac{i\eps + 1}{2}, & i < k',\\
            k'\cdot\eps, & i = k',\\
            i\eps \text{  or  } 1, & i > k'
        \end{cases}
        \quad\quad
        \textrm{and}
        \quad\quad
            \ell_{i}(\J_{k',t})
        =   \begin{cases}
            \frac{i\eps + 1}{2}, & i < k',\\
            1, & i = k',\\
            i\eps \text{  or  } 1, & i > k'.
        \end{cases}
        \]

        By our choice of $\eps = \frac{1}{4k}$, $\ell_{k'}$ is always the unique minimizer of $\ell(\I_{k',t})$. Thus, Algorithm~\ref{alg:ERM} always outputs $\ell_{k'}$. Moreover, when $k' \ne 1$ (which happens with probability $1 - \frac{1}{k} \ge \frac{1}{2}$), the resulting excess risk is at least
        $\ell_{k'}(\J_{k',t}) - \ell_{1}(\J_{k',t}) = 1 - \frac{\eps+1}{2} \ge \frac{1}{4}$.
        This proves the lower bound of $\frac{1}{8}$ on the expected excess risk incurred by Algorithm~\ref{alg:ERM}.
    \end{proof}


\bibliographystyle{alpha}
\bibliography{main}

\appendix

\section{Proof of Proposition~\ref{prop:selectivity}}\label{app:selectivity}
	\begin{proof}[Proof of Proposition~\ref{prop:selectivity}]
	    In the first case that $t$ is known to the adversary, we simply construct a binary sequence such that $x_{t+1} = \cdots = x_{n}$, and $x_{t+1}$ is randomly drawn from $\{0, 1\}$ with equal probability. When $\A$ makes a prediction at time $t$, the actual average of the sequence is either $0$ or $1$ with equal probability. It can be verified that any algorithm must achieve an expected squared loss of at least $\frac{1}{4} \ge \frac{1}{64}$.
	    
	    Now we consider the second case, where the window length $m$ is fixed. We choose $m' = \left\lceil\frac{m}{2}\right\rceil$ and construct a sequence consisting of blocks of length $m'$. Each block consists of the same value, which is chosen from $\{0, 1\}$ uniformly and independently at random. Whenever Algorithm $\A$ makes a prediction, by our choice of $m'$, the prediction window of size $m$ must contain an entire block. Since the variance in the average of the block is $\frac{1}{4}$ and the block contributes an $\frac{m'}{m}$ fraction to the average that $\A$ aims to predict, the variance in the arithmetic mean is then lower bounded by $\left(\frac{m'}{m}\right)^2 \cdot \frac{1}{4} \ge \frac{1}{64}$. This implies a lower bound of $\frac{1}{64}$ on the squared loss.
	    
	    In the third case,  the prediction algorithm chooses $t$, but is forced to make a prediction over the entire remaining $n-t$ timesteps.  In this case, consider constructing an adversarial distribution over sequences of length $n$ such that the first block of $n/2$ values are all identical and are chosen to either all be 0 or all be 1 with probability 1/2 of each choice, then next block of $n/4$ are identical and randomly selected to be either 0 or 1, and similarly for the next block of $n/8$, $n/16$, $n/32$, etc.  Let $t$ denote the time at which the prediction algorithm  makes its prediction.  There will always some $i$ for which the block of size $n/2^i$ is contained within the final $n-t$ timesteps, and for which $n/2^i$ is at least a $1/4$ fraction of $n-t$.  Hence the variance in the average value due to that block alone implies a lower bound of at least $\frac{1}{64}$ on the expected squared loss of any prediction.
	\end{proof}

\end{document}